\documentclass[11pt,letterpaper]{article}
\usepackage{naaclhlt2016}
\usepackage{times}
\usepackage{latexsym}
\usepackage{amssymb,amsfonts,amsmath}
\usepackage{graphicx}
\usepackage{multirow}
\usepackage{epsfig}
\usepackage{ascmac}
\usepackage{eclbkbox}

\naaclfinalcopy 
\def\naaclpaperid{***} 

\title{Analysis of the Effect of Dependency Information on Predicate-Argument Structure Analysis and Zero Anaphora Resolution}

\author{Koichiro Yoshino$^{1,3}$, Shinsuke Mori$^2$ and Satoshi Nakamura$^1$\\
	    $^1$Graduate School of Information Science, Nara Institute of Science and Technology\\
	    8916-5, Takayama-cho, Ikoma, Nara, 6300192, Japan\\
	    {\tt koichiro, s-nakamura@is.naist.jp}\\
	    $^2$Academic Center for Computing and Media Studies, Kyoto University\\
	    Yoshida-honmachi, Sakyo, Kyoto, 6068501, Japan\\
	    {\tt mori.shinsuke.8u@kyoto-u.ac.jp}\\
	    $^3$PRESTO, Japan Science and Technology Agency\\
	    4-1-8, Honmachi, Kawaguchi, Saitama, 3320012, Japan
	    }

\date{}

\begin{document}

\maketitle

\begin{abstract}
This paper investigates and analyzes the effect of dependency information on predicate-argument structure analysis (PASA) and zero anaphora resolution (ZAR) for Japanese, and shows that a straightforward approach of PASA and ZAR works effectively even if dependency information was not available.
We constructed an analyzer that directly predicts relationships of predicates and arguments with their semantic roles from a POS-tagged corpus. 
The features of the system are designed to compensate for the absence of syntactic information by using features used in dependency parsing as a reference.
We also constructed analyzers that use the oracle dependency and the real dependency parsing results, and compared with the system that does not use any syntactic information to verify that the improvement provided by dependencies is not crucial.
\end{abstract}

\section{Introduction}
\label{Section:Introduction}
Predicate-argument structure (PAS) including zero anaphora is one of the most fundamental and classical components of natural language processing (NLP).
Many NLP applications utilize PAS, such as machine translation \cite{zhai-zhang-zhou-zong:2013:ACL}, question answering \cite{shen-lapata:2007:EMNLP} and dialogue systems \cite{yoshino-mori-kawahara:2011:SIGDIAL}.

Conventionally, the PAS analysis (PASA) architecture stands on the pipeline processing of NLP.
It is assumed that the analyzer receives the correct results of various preprocessing steps such as word segments (WSs) and part of speech (POS) tags from a morphological analyzer, and dependency structures from a dependency parser.
However, underlying this assumption requires the costly processes of preparing an accurate morphological analyzer and dependency parser for its domain or application.
Actually the accuracies of dependency parsers are still not sufficient (around or less than 90\%) \cite{Kudo:2002:JDA:1118853.1118869,flannery-miyao-neubig-mori:2011:IJCNLP} to use as input of PASA even if the parser is adapted to the target domain.
This is in contrast to morphological analyzers adapted to the target domain \cite{neubig-nakata-mori:2011:ACL}, which have accuracies of more than 96\%.
Furthermore, the cost for constructing a dependency parser adapted to the target domain is much higher than the cost of construction of a domain adapted morphological analyzer, in both of data preparation and parser adaptation.

However, this approach still requires a domain adapted dependency parser.
In \cite{yoshino-mori-kawahara:2013:IJCNLP}, they proposed a straightforward framework that does not require any syntactic information, and directly predicts the pair of a predicate and an argument that has a relationship of semantic role from an entire document.
However, this work did not compare with a PAS analyzer that uses dependency information.
This paper follows our previous approach to construct a PAS analyzer that does not assume dependencies as input, compares it with a PAS analyzer that uses dependencies, and investigates the effect of dependency information.
This paper also studies influences of parsing errors and the cost of parser construction, which we often encounter in real language processing. 
The straightforward framework allows us to be free from costly processes of the data preparation and preprocessor construction if the accuracy is enough \cite{Zhou:2015:ACL}.

\section{Predicate Argument Structure Analysis (PASA)}
\label{Section:PASA}
\begin{figure*}[!t]
\centerline{\includegraphics[width=140mm,clip]{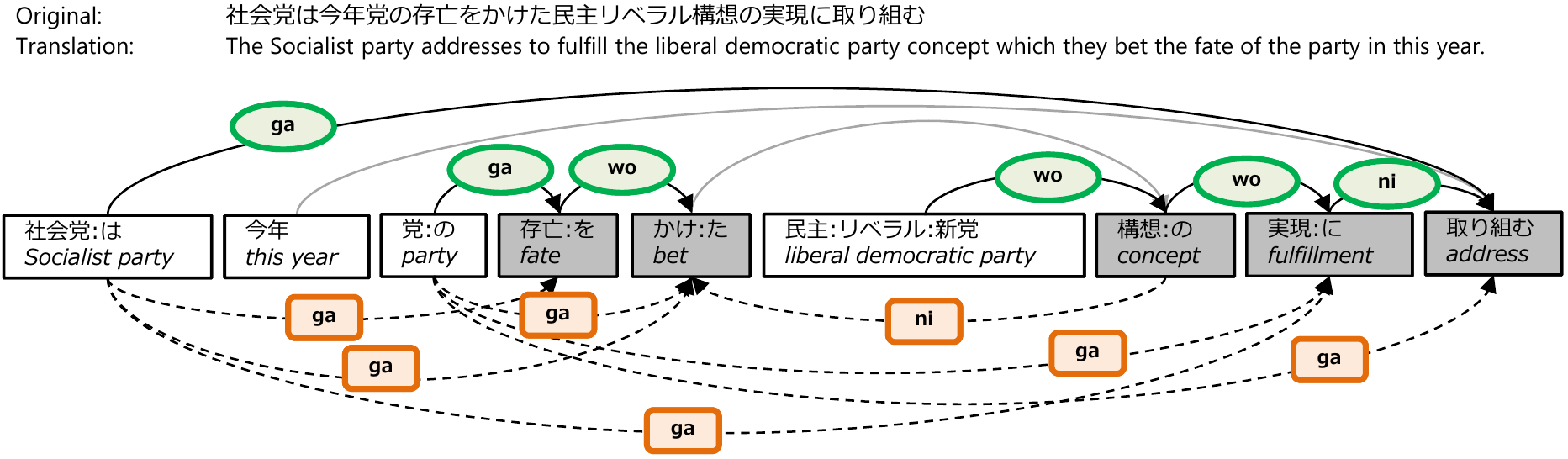}}
\caption{An example of the predicate-argument structure.
Arrows express relations in which leaves depend on the heads.
Solid lines mean dependencies, labels show their relations, and broken lines mean relations of the PAS that do not have dependency relations.
Gray lines are dependencies that do not have any PAS relation.
}
\label{Figure:PAS}
\end{figure*}

The PAS is a relationship between a predicate, a verb or an indeclinable word (noun, adjective, and adjective verbs) that indicates an event, and its arguments.
A predicate $P_i$ in a document $D$ has arguments $A_{i_1}$, $A_{i_2}$, ..., $A_{i_j}$ that have semantic roles $S_{i_1}$, $S_{i_2}$, ..., $S_{i_j}$.
We show an example of a PAS in {\bf Figure~\ref{Figure:PAS}}.
In the example, predicates $P_1$={\it fate}, $P_2$={\it bet}, $P_3$={\it concept}, $P_4$={\it fulfill}, and $P_5$={\it address} are given, and they have corresponding arguments $A_{1_{1}}$={\it party},...,$A_{5_{j}}$={\it fulfillment} with semantic roles $S_{1_{1}}$={\it ga},...,$S_{5_{j}}$={\it ni} (gray boxes are predicates).
Predicates can take the other predicates as their argument.
For example, $P_5$={\it address} takes {\it fulfillment} as its {\it ni}-case, although the {\it fulfillment} is also a predicate. 
The solid arrows (upper half of the figure) represent dependencies, and the leaves depend on the heads (dependency information does not include labels of edges).
Semantic role labels are annotated to the dependency edges (ellipsoidal labels).
Language dependent label values of these labels are defined in the shared task of semantic role labeling (SRL) \cite{Hajic:2009:CST:1596409.1596411}, respectively.
The example in the figure is a Japanese sentence.
In Japanese, the defined labels are {\it ga} (=nominative case), {\it wo} (=accusative case) and {\it ni} (=dative case).
Gray solids lines are dependencies that do not have defined semantic roles.
The dotted arrows (lower half of the figure) denote zero anaphora, which occur frequently in Japanese, and express predicate-argument relationships between words which do not have dependency relations.
The defined label-set is the same as the SRL label-set.
The task of predicting the zero anaphora in particular is called zero anaphora resolution (ZAR) \cite{iida2007zero,sasano09zero,iida2011cross}.
The example only shows zero anaphora relations between words in the same sentence, but zero anaphora relations also exist between words in different sentences.
ZAR is closely related to the coreference resolution task \cite{iida2003incorporating} that predicts identical elements in the real world written in a document.
If some words are identified as a coreference, each word of a coreferential cluster takes the same role of the same predicate.
In the example, the relation between the {\it Socialist party} and the {\it party} is a coreference, and both words take the {\it ga} label of predicates, {\it fate}, {\it bet}, {\it fulfillment}, and {\it address}.

The task of PASA is that of predicting arguments and their semantic roles for given predicates.
Conventional PAS analyzers output only one argument in one semantic role for a predicate and coreference resolution, post-processing, disambiguates the problem of multi arguments of a predicate in the same semantic role (e.g. both of {\it Socialist party} and {\it party} is {\it ga} (nominative case) argument of {\it address}) \cite{matsubayashi2012building}.
This paper follows this protocol, and defines the PASA task to output one argument in one semantic role for a predicate.
In other words, the task of PASA is predicting a part of a coreferential cluster that takes a semantic role for the given predicate.
In the example, for $P_5$={\it address}, the analyzer outputs one of {\it Socialist party} or {\it party} as the semantic role {\it ga} (nominative case).

\section{Pointwise Predicate-Argument Structure Analysis}
\label{Section:PWPASA}
Pointwise PASA (PWPASA) is a framework that predicts relations between every word in a document and a given predicate independently from other predicates with a binary classifier \cite{yoshino-mori-kawahara:2013:IJCNLP}.
One of the advantages of the pointwise approach is that the classifier does not depend on other tasks of NLP, whose prediction accuracy may not be sufficiently high \cite{neubig-nakata-mori:2011:ACL,mori11interspeech}.
The PWPASA does not essentially require global structures such as dependencies, and can be used to create a PASA model that does not refer to dependency information.

\subsection{Model of Prediction}
\label{Section:PWPASA:Prediction}
\begin{figure}[!t]
\centering
\includegraphics[width=75mm,clip]{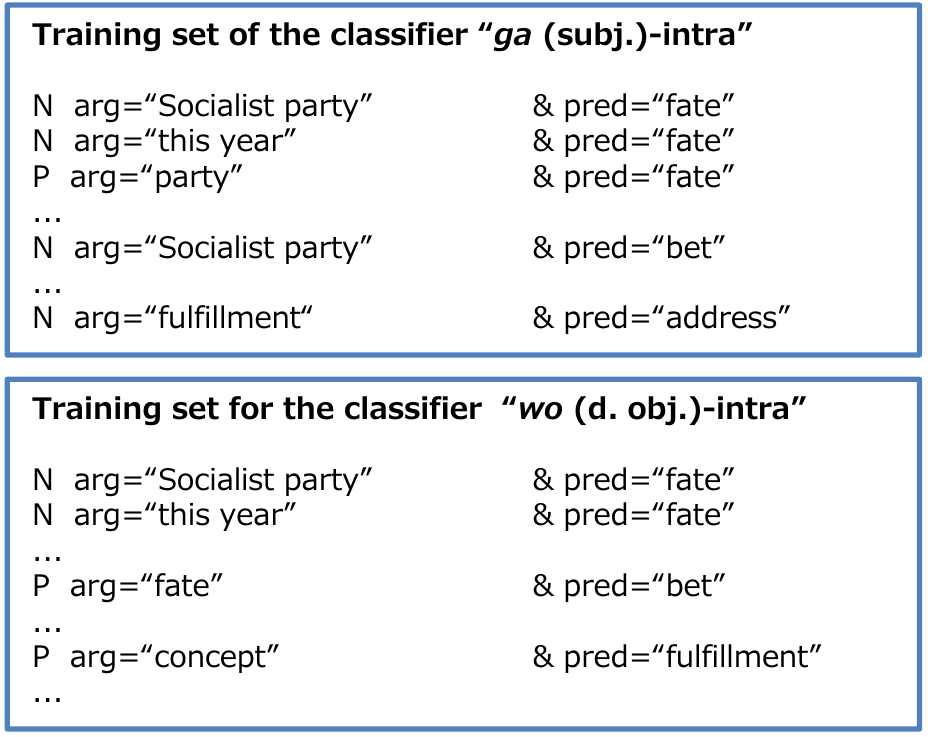}
\caption{Examples of positive and negative example creation from the example of Figure \ref{Figure:PAS}.
This figure shows examples of the label {\it ga} in the {\it intra} case and the label {\it wo} in the {\it intra} case.}
\label{Figure:PWPASA}
\end{figure}

PWPASA handles the problem of PASA as a binary classification problem for every pair of an argument candidate and a predicate.
Labeled pairs of an argument and a predicate are used as positive training examples ({\sf P}) and unlabeled pairs are used as negative training examples ({\sf N}).
In the example of {\bf Figure \ref{Figure:PWPASA}}, the upper box shows training examples for the {\it ga}-{\it intra} classifier.
The pair of ``fate'' and ``party'', and the pair of ``fate'' and ``Socialist party'' are positive examples, and pairs of ``fate'' and other candidates are negative.
Similarly the bottom box shows some training examples for the {\it wo}-{\it intra} classifier converted from the example in Figure~\ref{Figure:PAS}.
The system consists of 3$\times$2$=$6 classifiers in total: {\it ga} (nominative case), {\it wo} (accusative case) and {\it ni} (dative case) for intra-sentence ({\it intra}) or inter-sentence ({\it inter}) cases.
The {\it intra} classifiers are trained from all pairs of a predicate and every word in the same sentence, and {\it inter} ones are trained from all pairs of predicates and every word in different sentences.

\begin{figure}[!t]
\centering
\includegraphics[width=75mm,clip]{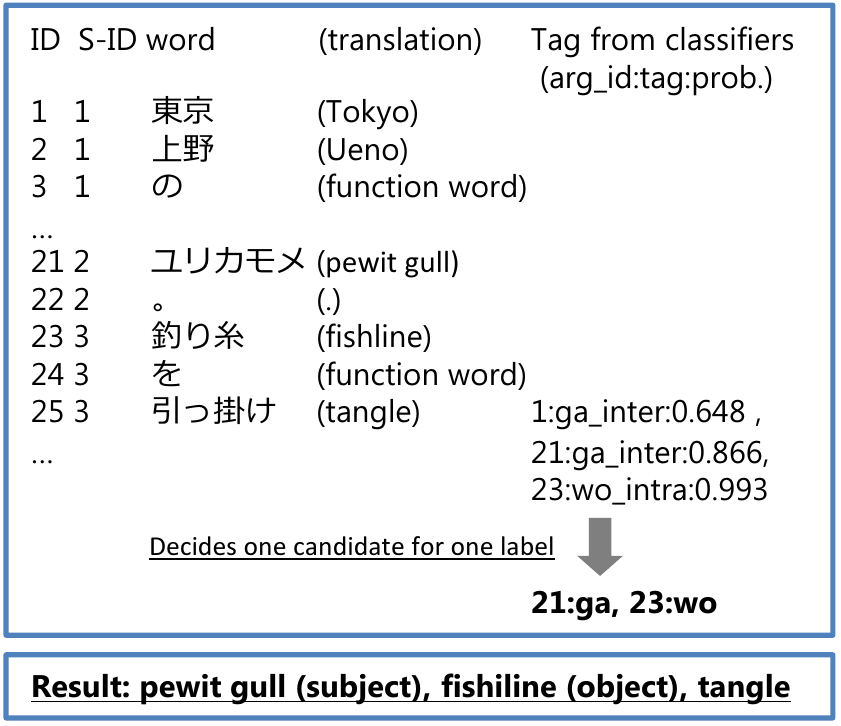}
\caption{An example of classification of several classifiers and a candidate decision.}
\label{Figure:Prediction}
\end{figure}

The PASA task expects to output one argument for one predicate in one semantic role.
However, the PWPASA outputs several argument candidates for a predicate in one semantic role, thus, we used a logistic regression (LR) classifier to select the best candidate and its prediction probability.
As shown in {\bf Figure~\ref{Figure:Prediction}}, classifiers for the same semantic role output several classification results.
Here, {\sf ID} is word ID and {\sf S-ID} is sentence ID.
The classification results sometimes conflict because the labeling for each pair of an argument candidate and a predicate is independent. 
(``Tokyo'' and ``pewit gull'' are classified into nominatives of the predicate ``tangle'' in the example of {\bf Figure~\ref{Figure:Prediction}}).
The system outputs one result that has the highest probability of the LR classifiers for one semantic role to a predicate.
In the example shown in Figure~\ref{Figure:Prediction}, the results of ``Tokyo'' and ``pewit gull'' for ``tangle'' conflict.
In this case, the system chooses one of them for the nominative case with the prediction probabilities of LR.

\newcommand{\onesp}{\hspace{1.0mm}}

\begin{figure*}[!t]
\centering
\includegraphics[width=140mm,clip]{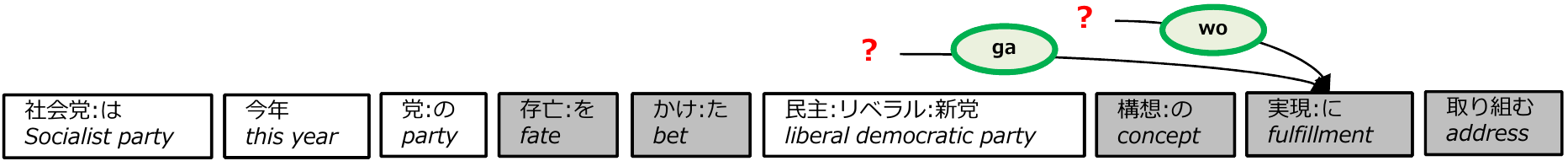}
\caption{An example of gold case frame.}
\label{Figure:GCF}
\end{figure*}

\subsection{Features for PWPASA ({\sf PWFeat})}
\label{Section:PWPASA:Feature}
Feature design is the most important aspect of the PWPASA.
The classifier does not use information given by dependencies, but it indirectly refers to the information by using features that reflect dependency relationships.
In the subsequent explanations, features indicated by $\dagger$ are ones generally used in the dependency parsing \cite{flannery-miyao-neubig-mori:2011:IJCNLP}.
\begin{enumerate}
  \setlength{\parskip}{0cm} 
  \setlength{\itemsep}{0cm} 
\item {\bf $^\dagger$Word {\it n}-gram}: 
Uni-grams of words located between -10 -- and +10 positions around the predicate ($w_{p-10}$--$w_{p+10}$) and the argument candidate ($w_{a-10}$--$w_{a+10}$),
bi-grams and tri-grams of words located -5 -- +5 around the predicate and the argument ($w_{p-5}w_{p-4}$, ..., $w_{p+4}w_{p+5}$, $w_{a-5}w_{a-4}$, ..., $w_{a+4}w_{a+5}$, 
$w_{p-5}w_{p-4}w_{p-3}$, ..., $w_{p+3}w_{p+4}w_{p+5}$ and $w_{a-5}w_{a-4}w_{a-3}$, ..., $w_{a+3}w_{a+4}w_{a+5}$).
\item {\bf $^\dagger$POS {\it n}-gram}:
Uni-grams, bi-grams and tri-grams of POS tags of the surrounding words. 

\item {\bf $^\dagger$Pairwise word and POS}:
The pair of the word of the predicate and the argument candidate ($w_{p+0}w_{a+0}$) and pairs of POS tags of surrounding words located -2 -- +2 ($t(w_{p-2})t(w_{a-2})$, $t(w_{p-2})t(w_{a-1})$, ..., $t(w_{p+2})t(w_{a+2})$).

\item {\bf $^\dagger$Word Distance}:
The number of words between the predicate and the argument candidate. 
The number as is, and divided and rounded values with 2, 3, 4 and 5.
The word distance features take integer values which include negative integers, to make distinction of right and left of the predicate.

\item {\bf Predicate Distance}:
\label{Section:Feature:PDist}
The number of predicates between the predicate and the candidate. 

\item {\bf Gold case frame}:
The gold case frame of the target predicate.
This information is similar to the results of the word sense disambiguation.
\nocite{watanabe10aclshort}
An example is shown in {\bf Figure~\ref{Figure:GCF}}.
If the word sense disambiguation worked perfectly, the system knows that the {\it ga} and the {\it wo} label is essential for the predicate ``fulfill'', and these arguments exist somewhere in the document.
On the other hand, system can know that it is not necessary to find an argument that takes the role of {\it ni}.
This feature is introduced as a binary flag for every semantic role.

\end{enumerate}

\subsection{Features for PWPASA that Depend on Language ({\sf Lang})}
\label{Section:PWPASA:Langfeature}
The other important viewpoint is language dependence.
We list language dependent features of Japanese PASA.

\begin{enumerate}
  \setlength{\parskip}{0cm} 
  \setlength{\itemsep}{0cm} 
\item {\bf Case Marker Word on the Right Side}:
\label{Section:Feature:FW}
Case marker words ��~({\it ha}), ��~({\it ga}), ��~({\it wo}) and ��~({\it ni}) on the right side. 

\item {\bf Candidate Position}:
\label{Section:Feature:Bin}
The candidate position in a document is used as a feature.
\nocite{Grosz:1995:CFM:211190.211198}

\item {\bf Case Marker Word Distance}:
Number of case markers words between the predicate and the argument candidate.
 
\item {\bf Pair of Predicate Distance and Case Marker Word Distance}:
Pairwise features of the predicate distance and the case marker word distance. 
\end{enumerate}

\begin{figure*}[!t]
\centering
\includegraphics[width=140mm,clip]{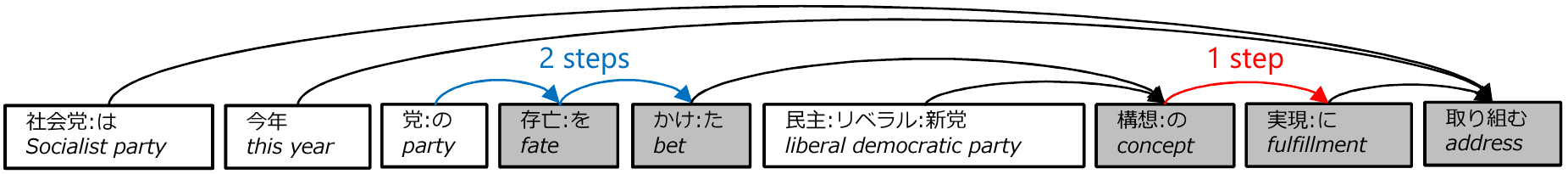}
\caption{An example of depth of dependency.}
\label{Figure:Depth}
\end{figure*}

\subsection{Features of Dependency Information ({\sf Dep})}
\label{Section:PWPASA:Depfeature}
This section describes features referring to dependency information to compare with the pointwise analyzer (PWPASA).
PASA with dependency introduces two kinds of dependency features generally used for PASA.
We selected features which improve the SRL accuracy significantly in a previous study \cite{Bjorkelund:2009:MSR:1596409.1596416}.

\begin{enumerate}
  \setlength{\parskip}{0cm} 
  \setlength{\itemsep}{0cm} 
\item {\bf Dependency Relation between Predicate and Argument Candidate}:
The dependency relation between the predicate and the argument candidate. 
We introduced the direct dependency relation feature for distance of up to 3 steps.
{\bf Figure~\ref{Figure:Depth}} shows examples of this feature.

\item {\bf Head and Leaf Words}:
Head word of the predicate ($w_{p_h}$), the head word of the argument candidate ($w_{a_h}$), and the leaf words of the predicate ($w_{p_{l_1}}$, ...,$w_{p_{l_i}}$) and the leaf words of the argument candidate ($w_{a_{l_1}}$, ...,$w_{a_{l_j}}$).
For example in Figure~\ref{Figure:Depth}, the head of ``concept'' is ``fulfillment'', and leafs of it are ``bet'' and ``liberal democratic party''.
\end{enumerate}

\section{Experimental Evaluation}
\label{Section:Eval}
We investigated the effect of the dependency information on PASA including ZAR by comparing the proposed analyzer which does not use the dependency information and a more traditional system that uses dependency information. 
All of experimental conditions except for the dependency information are the same.
We constructed two systems that do not use dependency information; {\sf PWFeat} and {\sf PWFeat+Lang}, and three systems that use dependency information; oracle dependency ({\sf PWFeat+Lang+Dep(oracle)}), real parsing results ({\sf PWFeat+Lang+Dep(parsed)}), and simulated parsing results with 20\% errors ({\sf PWFeat+Lang+Dep(20\% errors)}).
Used feature type is indicated in names, {\sf PWFeat} means using features for PWPASA, {\sf Lang} means using language dependent features, and {\sf Dep} means using features of dependency information.
For {\sf parsed}, we used a phrase-based dependency parser CaboCha \cite{Kudo:2002:JDA:1118853.1118869}, which is retrained by the following training set of the PASA.
The parsing accuracy was 86.12\% (the parsing result includes 13.88\% errors).

\subsection{Experimental Setting}
We used the NAIST Text Corpus (NTC) \cite{Iida:2007:AJT:1642059.1642081}, a corpus which is publicly available\footnote{https://sites.google.com/site/naisttextcorpus (2015/5/30)}.
The documents are annotated with predicate-argument relations and coreferences.
The sentences also have lower layer annotations: word boundaries, POS tags, chunks, and phrase-based dependencies.
The annotated predicates include not only verbs but also indeclinable words which indicate events.
The NTC contains Japanese newspaper articles and editorials.
There are three different types of annotation on pairs of a predicate and its argument in Japanese: {\it ga} (nominative case), {\it wo} (accusative case) and {\it ni} (dative case).

\begin{table}[t]
\caption{\label{Table:Corpus} Detail of training and test set.}
\begin{center}
\scriptsize{
\begin{tabular}{|@{\onesp}l|@{\onesp}l@{\onesp}|@{\onesp}l|r@{\onesp}|r@{\onesp}|}
\hline
\multicolumn{3}{|l|}{}																& Train 		& Test\\
\hline
\hline
\multicolumn{3}{|l|}{Documents} 											& 1,751 		& 696\\
\hline
\multicolumn{3}{|l|}{Sentences}												& 24,283		& 9,284\\
\hline
\multicolumn{3}{|l|}{Words}													& 664,898 	& 225,624\\
\hline
\multicolumn{3}{|l|}{Predicates} 											& 97,773 	& 38,365\\
\hline
\multirow{9}{*}{PAS labels}	& {\it ga}	& Depend				& 64,152 	& 12,226\\
\cline{3-5}
													& (nom.)	& Zero~({\it intra})		& 62,586 	& 14,373\\
\cline{3-5}
													&  				& Zero~({\it inter})		& 149,482 	& 49,415\\
\cline{2-5}
													& {\it wo}	& Depend				& 51,095 	& 9,837\\
\cline{3-5}
													& (acc.)		& Zero~({\it intra})		& 17,585 	& 3,179\\
\cline{3-5}
													&  				& Zero~({\it inter})		& 10,786 	& 3,830\\
\cline{2-5}
													& {\it ni}		& Depend				& 11,790 	& 2,501\\
\cline{3-5}
													& (dat.)		& Zero~({\it intra})		& 4,063 		& 1,005\\
\cline{3-5}
													&  				& Zero~({\it inter})		& 6,978 		& 2,048\\
\hline
\end{tabular}
}
\end{center}
\end{table}

We divided the NTC into training and test set.
Their specifications are shown in {\bf Table~\ref{Table:Corpus}}.
The table shows the numbers of documents, sentences, words, predicates and PAS labels annotated on pairs of a predicate and its argument.
The PAS label numbers include several labels that indicate the same coreferential cluster from the same predicate in the same semantic role (In the example of Figure~\ref{Figure:PAS}, there are two arcs with a {\it ga}~(nominative case) labels from ``fulfillment'' to ``Socialist party'' and ``party'', and ``Socialist party'' and ``party'' belong to the same coreferential cluster).
As the classifier, we used LIBLINEAR\footnote{http://www.csie.ntu.edu.tw/~cjlin/liblinear (2015/5/30)}, a library for large linear classification \cite{Fan:2008:LLL:1390681.1442794}, with L2-loss linear logistic regression (LR).
We evaluated the system performance by using precision (P), recall (R), and their harmonic mean (F-measure; F).

\begin{table}[t]
\caption{\label{Table:ALL}Overall result of each method.}
\begin{center}
\scriptsize{
\begin{tabular}{|l|l|@{\onesp}r@{\onesp}|@{\onesp}r@{\onesp}|@{\onesp}r@{\onesp}|}
\hline
Method & Category & P & R & F\\
\hline
\hline
{\sf PWFeat} & Depend & 93.01 & 69.20 & 79.36 \\
\cline{2-5}
& Zero~({\it intra}) & 51.05 & 23.96 & 32.61 \\
\cline{2-5}
& Zero~({\it inter}) & 23.55 & 5.00 & 8.25 \\
\cline{2-5}
& ALL & 82.57 & 50.25 & 62.48 \\
\hline
{\sf PWFeat+Lang} & Depend & 94.00 & 70.48 & 80.56 \\
\cline{2-5}
& Zero~({\it intra}) & 52.97 & 24.24 & 33.26 \\
\cline{2-5}
& Zero~({\it inter}) & 25.00 & 5.83 & 9.46 \\
\cline{2-5}
& ALL & 83.64 & 51.25 & {\bf 63.56} \\
\hline
{\sf PWFeat+Lang} & Depend & 92.64 & 76.93 & 84.06 \\
\cline{2-5}
{\sf +Dep(oracle)} & Zero~({\it intra}) & 58.72 & 19.11 & 28.84 \\
\cline{2-5}
& Zero~({\it inter}) & 23.10 & 5.65 & 9.08 \\
\cline{2-5}
& ALL & 85.09 & 54.26 & {\bf 66.26} \\
\hline
{\sf PWFeat+Lang} & Depend & 92.82 & 75.06 & 83.00 \\
\cline{2-5}
{\sf +Dep(parsed)}& Zero~({\it intra}) & 57.45 & 21.72 & 31.52 \\
\cline{2-5}
& Zero~({\it inter}) & 23.79 & 5.83 & 9.37 \\
\cline{2-5}
& ALL & 84.47 & 53.64 & {\bf 65.61} \\
\hline
{\sf PWFeat+Lang} & Depend & 92.65 & 72.22 & 81.17 \\
\cline{2-5}
{\sf +Dep(20\% errors)}& Zero~({\it intra}) & 54.57 & 18.93 & 28.11 \\
\cline{2-5}
& Zero~({\it inter}) & 23.88 & 5.31 & 8.69 \\
\cline{2-5}
& ALL & 84.34 & 51.14 & {\bf 63.67} \\
\hline
\end{tabular}
}
\end{center}
\end{table}

\begin{table}[t]
\caption{\label{Table:Proposed}{\sf PWPFeat+Lang}: Results of the analyzer that does not use the dependency information.}
\begin{center}
\scriptsize{
\begin{tabular}{|l|l|@{\onesp}r@{\onesp}|@{\onesp}r@{\onesp}|@{\onesp}r@{\onesp}|}
\hline
Role type & Category & P & R & F\\
\hline
\hline
\multirow{8}{*}{{\it ga} (nom.)} & \multirow{2}{*}{Depend} & 90.16 & 60.22 & 72.21 \\
& & (9516/10554) & (9516/15803) & \\
\cline{2-5}
& \multirow{2}{*}{Zero~({\it intra})} & 52.75 & 24.75 & 33.69 \\
& & (1957/3710) & (1957/7906) & \\
\cline{2-5}
& \multirow{2}{*}{Zero~({\it inter})} & 24.76 & 6.56 & 10.37 \\
& & (354/1430) & (354/5396) & \\
\cline{2-5}
& \multirow{2}{*}{ALL} & 75.36 & 40.64 & 52.80 \\
& & (11827/15694) & (11827/29105) & \\
\hline
\multirow{8}{*}{{\it wo} (acc.)} & \multirow{2}{*}{Depend} & 97.65 & 82.70 & 89.56 \\
& & (8751/8962) & (8751/10581) & \\
\cline{2-5}
& \multirow{2}{*}{Zero~({\it intra})} & 59.48 & 22.68 & 32.84 \\
& & (320/538) & (320/1411) & \\
\cline{2-5}
& \multirow{2}{*}{Zero~({\it inter})} & 36.17 & 2.02 & 3.82 \\
& & (17/47) & (17/843) & \\
\cline{2-5}
& \multirow{2}{*}{ALL} & 95.19 & 70.81 & 81.21 \\
& & (9088/9547) & (9088/12835) & \\
\hline
\multirow{8}{*}{{\it ni} (dat.)} & \multirow{2}{*}{Depend} & 97.47 & 82.91 & 89.60 \\
& & (2193/2250) & (2193/2645) & \\
\cline{2-5}
& \multirow{2}{*}{Zero~({\it intra})} & 40.19 & 19.76 & 26.50 \\
& & (84/209) & (84/425) & \\
\cline{2-5}
& \multirow{2}{*}{Zero~({\it inter})} & 20.51 & 3.10 & 5.39 \\
& & (8/39) & (8/258) & \\
\cline{2-5}
& \multirow{2}{*}{ALL} & 91.47 & 68.66 & 78.44 \\
& & (2285/2498) & (2285/3328) & \\
\hline
\multirow{8}{*}{ALL} & \multirow{2}{*}{Depend} & 94.00 & 70.48 & 80.56 \\
& & (20460/21766) & (20460/29029) & \\
\cline{2-5}
& \multirow{2}{*}{Zero~({\it intra})} & 52.97 & 24.24 & 33.26 \\
& & (2361/4457) & (2361/9742) & \\
\cline{2-5}
& \multirow{2}{*}{Zero~({\it inter})} & 25.00 & 5.83 & 9.46 \\
& & (379/1516) & (379/6497) & \\
\cline{2-5}
& \multirow{2}{*}{ALL} & 83.64 & 51.25 & 63.56 \\
& & (23200/27739) & (23200/45268) & \\
\hline
\end{tabular}
}
\end{center}
\end{table}

\begin{table}[t]
\caption{\label{Table:Dependency}{\sf PWFeat+Lang+Dep(oracle)}: Results of the analyzer that uses the gold dependency information.}
\begin{center}
\scriptsize{
\begin{tabular}{|l|l|@{\onesp}r@{\onesp}|@{\onesp}r@{\onesp}|@{\onesp}r@{\onesp}|}
\hline
Role type & Category & P & R & F\\
\hline
\hline
\multirow{8}{*}{{\it ga} (nom.)} & \multirow{2}{*}{Depend} & 88.67 & 70.20 & 78.36 \\
& & (11093/12511) & (11093/15803) & \\
\cline{2-5}
& \multirow{2}{*}{Zero~({\it intra})} & 58.62 & 19.10 & 28.81 \\
& & (1510/2576) & (1510/7906) & \\
\cline{2-5}
& \multirow{2}{*}{Zero~({\it inter})} & 22.99 & 6.36 & 9.96 \\
& & (343/1492) & (343/5396) & \\
\cline{2-5}
& \multirow{2}{*}{ALL} & 78.09 & 44.48 & 56.68 \\
& & (12946/16579) & (12946/29105) & \\
\hline
\multirow{8}{*}{{\it wo} (acc.)} & \multirow{2}{*}{Depend} & 97.01 & 85.00 & 90.61 \\
& & (8994/9271) & (8994/10581) & \\
\cline{2-5}
& \multirow{2}{*}{Zero~({\it intra})} & 65.33 & 19.63 & 30.19 \\
& & (277/424) & (277/1411) & \\
\cline{2-5}
& \multirow{2}{*}{Zero~({\it inter})} & 29.63 & 1.90 & 3.57 \\
& & (16/54) & (16/843) & \\
\cline{2-5}
& \multirow{2}{*}{ALL} & 95.26 & 72.36 & 82.25 \\
& & (9287/9749) & (9287/12835) & \\
\hline
\multirow{8}{*}{{\it ni} (dat.)} & \multirow{2}{*}{Depend} & 96.64 & 84.88 & 90.38 \\
& & (2245/2323) & (2245/2645) & \\
\cline{2-5}
& \multirow{2}{*}{Zero~({\it intra})} & 43.86 & 17.65 & 25.17 \\
& & (75/171) & (75/425) & \\
\cline{2-5}
& \multirow{2}{*}{Zero~({\it inter})} & 18.60 & 3.10 & 5.32 \\
& & (8/43) & (8/258) & \\
\cline{2-5}
& \multirow{2}{*}{ALL} & 91.76 & 69.95 & 79.38 \\
& & (2328/2537) & (2328/3328) & \\
\hline
\multirow{8}{*}{ALL} & \multirow{2}{*}{Depend} & 92.64 & 76.93 & 84.06 \\
& & (22332/24105) & (22332/29029) & \\
\cline{2-5}
& \multirow{2}{*}{Zero~({\it intra})} & 58.72 & 19.11 & 28.84 \\
& & (1862/3171) & (1862/9742) & \\
\cline{2-5}
& \multirow{2}{*}{Zero~({\it inter})} & 23.10 & 5.65 & 9.08 \\
& & (367/1589) & (367/6497) & \\
\cline{2-5}
& \multirow{2}{*}{ALL} & 85.09 & 54.26 & 66.26 \\
& & (24561/28865) & (24561/45268) & \\
\hline
\end{tabular}
}
\end{center}
\end{table}

\begin{table}[t]
\caption{\label{Table:Dependency-parse}{\sf PWFeat+Lang+Dep(parsed)}: Results of the analyzer that uses the result of the parser as the dependency information (features are the same as {\sf PWFeat+Lang+Dep(oracle)}).}
\begin{center}
\scriptsize{
\begin{tabular}{|l|l|@{\onesp}r@{\onesp}|@{\onesp}r@{\onesp}|@{\onesp}r@{\onesp}|}
\hline
Role type & Category & P & R & F\\
\hline
\hline
\multirow{8}{*}{{\it ga} (nom.)} & \multirow{2}{*}{Depend} & 88.89 & 67.45 & 76.70 \\
& & (10659/11991) & (10659/15803) & \\
\cline{2-5}
& \multirow{2}{*}{Zero~({\it intra})} & 57.57 & 21.88 & 31.71 \\
& & (1730/3005) & (1730/7906) & \\
\cline{2-5}
& \multirow{2}{*}{Zero~({\it inter})} & 23.70 & 6.60 & 10.32 \\
& & (356/1502) & (356/5396) & \\
\cline{2-5}
& \multirow{2}{*}{ALL} & 77.25 & 43.79 & 55.90 \\
& & (12745/16498) & (12745/29105) & \\
\hline
\multirow{8}{*}{{\it wo} (acc.)} & \multirow{2}{*}{Depend} & 97.05 & 84.22 & 90.18 \\
& & (8911/9182) & (8911/10581) & \\
\cline{2-5}
& \multirow{2}{*}{Zero~({\it intra})} & 63.49 & 21.69 & 32.33 \\
& & (306/482) & (306/1411) & \\
\cline{2-5}
& \multirow{2}{*}{Zero~({\it inter})} & 31.37 & 1.90 & 3.58 \\
& & (16/51) & (16/843) & \\
\cline{2-5}
& \multirow{2}{*}{ALL} & 95.04 & 71.94 & 81.89 \\
& & (9233/9715) & (9233/12835) & \\
\hline
\multirow{8}{*}{{\it ni} (dat.)} & \multirow{2}{*}{Depend} & 96.44 & 83.89 & 89.73 \\
& & (2219/2301) & (2219/2645) & \\
\cline{2-5}
& \multirow{2}{*}{Zero~({\it intra})} & 40.82 & 18.82 & 25.76 \\
& & (80/196) & (80/425) & \\
\cline{2-5}
& \multirow{2}{*}{Zero~({\it inter})} & 17.50 & 2.71 & 4.70 \\
& & (7/40) & (7/258) & \\
\cline{2-5}
& \multirow{2}{*}{ALL} & 90.89 & 69.29 & 78.64 \\
& & (2306/2537) & (2306/3328) & \\
\hline
\multirow{8}{*}{ALL} & \multirow{2}{*}{Depend} & 92.82 & 75.06 & 83.00 \\
& & (21789/23474) & (21789/29029) & \\
\cline{2-5}
& \multirow{2}{*}{Zero~({\it intra})} & 57.45 & 21.72 & 31.52 \\
& & (2116/3683) & (2116/9742) & \\
\cline{2-5}
& \multirow{2}{*}{Zero~({\it inter})} & 23.79 & 5.83 & 9.37 \\
& & (379/1593) & (379/6497) & \\
\cline{2-5}
& \multirow{2}{*}{ALL} & 84.47 & 53.64 & 65.61 \\
& & (24284/28750) & (24284/45268) & \\
\hline
\end{tabular}
}
\end{center}
\end{table}

\subsection{Effect of Dependency Information}
\label{Section:Eval:Dep}
{\bf Table~\ref{Table:ALL}} shows a summary in several settings.
{\sf PWFeat+Lang} is the result of the proposed PASA analyzer that does not use dependency information.
Dependency features ({\sf Dep}) are additionally used in different settings: {\sf oracle} (gold dependencies), {\sf parsed} (real parsing results including errors), and {\sf 20\% errors} (simulated parsing results including 20\% errors).
{\bf Table~\ref{Table:Proposed}} shows the detailed results of the PWPASA analyzer that does not use the dependency information ({\sf PWFeat+Lang}), and {\bf Table~\ref{Table:Dependency}} and {\bf Table~\ref{Table:Dependency-parse}} show the results of the analyzers that additionally use the dependency information ({\sf PWFeat+Lang+Dep(oracle)} and {\sf PWFeat+Lang+Dep(parsed)}).
Role type is the type of PAS labels ({\it ga}, {\it wo}, {\it ni} in Japanese), category is the relation type of the argument and the predicate.
``Depend'' means an element of the coreferential cluster of the argument have a dependency relation to the predicate.
``Zero~{\it intra}'' means that one or more elements of the coreferential cluster exist in the same sentence as the predicate, but they do not depend on the predicate.
``Zero~{\it inter}'' does not include any element that exists in the same sentence of the predicate.

In total accuracy (ALL-ALL), the F-measure of {\sf PWFeat+Lang} was 63.56, 2.70 points lower than {\sf PWFeat+Lang+Dep(oracle)}.
The difference is statistically significant (p$<$0.01), however, it requires the oracle dependency information.
If the dependency was a real parsing result including errors ({\sf PWFeat+Lang+Dep(parsed)}), the difference of F-measure compared to the {\sf PWFeat+Lang} is much smaller, 2.05 points.
This indicates that the dependency features is still effective even if some features used in dependency parsing are indirectly used for the PAS analyzer,
however, the accuracy of {\sf PWFeat+Lang} closes in upon {\sf PWFeat+Lang+Dep(oracle)}.
These results also indicate that the parsing errors do not fatally influence the PASA accuracy, if the parser is trained in the same domain.
On the other hand, the dependency parser requires annotations of 24,283 sentences for the 2.05 points improvement.
In our other experiment, annotation speed of dependency was 22.40 sentences (consisting of 24.76 words) per hour (74,865[sent.]$/$135[hour]), in other words, 509.76 words per hour.
Preparing dependency information is an indirect method for domain adaptation of PASA, and the data preparation cost is very high.
The annotation of dependencies is very difficult for untrained annotators, and it requires spending a lot of time and energy of the developers.


If the parser is not adapted to the target domain, the parsing error rate reaches up to 20\% \cite{flannery-miyao-neubig-mori:2011:IJCNLP}.
We simulated parsing errors of the test set by randomly replacing a predefined proportion of the edges of the gold data of dependency, and used it on the test set ({\sf PWFeat+Lang+Dep(20\% errors)}).
The total accuracy (ALL-ALL) of the {\sf PWFeat+Lang} model was comparable to the result of the 20\% noisy test-set.
These results show that the result of a dependency parser achieves this low level of accuracy, because it is not adapted to the domains is not helpful for PASA.

In detailed categories, we compared the analyzer that does not use the dependency information ({\sf PWFeat+Lang}) and the analyzer that uses real parsing results ({\sf PWFeat+Lang+Dep(parsed)}).
The F-measure of the nominative cases that directly depend on predicates ({\it ga}~(nom.)-Depend) was mainly decreased by the absence of the dependency information (from 76.70 points to 72.21 points).
On the other hand, F-measures of the accusative and the dative cases, which depend on predicates ({\it wo}~(acc.)-Depend and {\it ni}~(dat.)-Depend), {\sf PWFeat+Lang} performs comparably to the analyzer that uses the dependency information.
These results show that the dependency information contributes to predict nominative cases that exist at more distant positions from predicates than accusative cases, especially in SOV languages.

In category of Zeros (Zero~({\it intra}) and Zero~({\it inter}); arguments which do not depend on their predicates, the dependency information negatively influenced all categories.
In the Zero~{\it intra} category (see ALL-Zero~({\it intra}) of Table~\ref{Table:Proposed}, Table~\ref{Table:Dependency} and Table~\ref{Table:Dependency-parse}), the F-measure of the {\sf PWFeat+Lang} model was 1.74 points (33.26-31.52) better than the F-measure of the {\sf PWFeat+Lang+Dep(parsed)} model, and 4.42 points (33.26-28.84) better than the F-measure of the {\sf PWFeat+Lang+Dep(oracle)} with a statistical significance (p$<$0.01).
In Zero~{\it inter} category (see ALL-Zero~({\it inter}) of Table~\ref{Table:Proposed}, Table~\ref{Table:Dependency} and Table~\ref{Table:Dependency-parse}), the difference was not significant but the F-measure of the {\sf PWFeat+Lang} model was also better than F-measures of the {\sf PWFeat+Lang+Dep(oracle)} and the {\sf PWFeat+Lang+Dep(parsed)}.
This is caused by the fact that the dependency information gives a strong prior to select an argument candidate that depends on the predicate, even if the candidate is not an argument.

\subsection{Effect of language specific features}
\label{Section:Eval:Dependence}
We also prepared the analyzer that does not use language specific features of the PWPASA (features described in Section~\ref{Section:PWPASA:Langfeature}) to verify the generality of the constructed {\sf PWFeat+Lang}.
By comparing to {\sf PWFeat} and {\sf PWFeat+Lang} in Table~\ref{Table:ALL}, 
the total accuracy (ALL-ALL) was decreased by 1.08 points by the lack of the language specific features.
It shows that there is a certain level of generality of the {\sf PWFeat+Lang} model that covers the dependency information with indirect features.

\section{Related Studies}
In the CoNLL shared task 2004 \cite{Carreras:2004:ICS:1706543.1706571,hacioglu2004semantic}, some systems tackled SRL only with shallow syntax structures; chunks and clauses.
In CoNLL shared task 2005 \cite{Carreras:2005:ICS:1706543.1706571,Koomen:2005:GIM:1706543.1706576}, more deep syntax structures, dependencies, are provided, and the best system outperformed the best system of CoNLL shared task 2004.
However, effects of parsing errors are not deeply analyzed, and these tasks only target labeling semantic roles between elements in the same sentences.

While applications that use PAS demand higher accuracy of PASA in various domains \cite{imamura-higashinaka-izumi:2014:COLING}, however, it is difficult to improve the accuracy of PASA on a variety of domains because of the lack of annotated corpora in each domain.
Constructing new annotated corpora in the new domains is very costly, thus, some previous studies tried to ignore the problem.
Semi-supervised and unsupervised approaches \cite{furstenau-lapata:2009:EACL,titov-klementiev:2012:PAPERS,lorenzo-cerisara:2014:CLITP} tried to improve the accuracy of labeling with unlabeled data, but they did not provide much of an improvement from the unlabeled data.
\cite{ribeyre:2015:NAACL} also researched the effect of syntax information for English SRL, and reported a 2.80 point improvement of accuracy.
Our work focuses on not only SRL but also ZAR.

Having training data that are representative of the domain is essential for constructing a robust PAS analyzer \cite{Pradhan:2008:TRS:1403157.1403163}.
However, data annotation of the higher-layer NLP tasks such as PASA require not only the annotations for the task but also those of lower-layer NLP task, such as dependencies. 
This property makes it difficult to apply the current supervised approaches to a new domain.
Our work tackled this problem by avoiding to analyzing the dependency structure.

In the PASA and the ZAR of Japanese, \cite{Imamura:2009:DAP:1667583.1667611} addressed PASA and ZAR with a discriminative model for only verbal predicates.
The total accuracy of our work is less than their work, but our system addressed both PASA and ZAR not only for verbal predicates but also for indeclinable predicates which indicate events.
Our system also worked better in ZAR of accusative and dative cases.
\cite{weko_79538_1} proposed a discriminative ZAR model.
Our system worked better than their work in the accusative case ({\it wo}) and the dative case ({\it ni}), but their system worked better in the nominative case ({\it ga}).

\begin{figure}[!t]
\centering
\includegraphics[angle=90, width=75mm, clip]{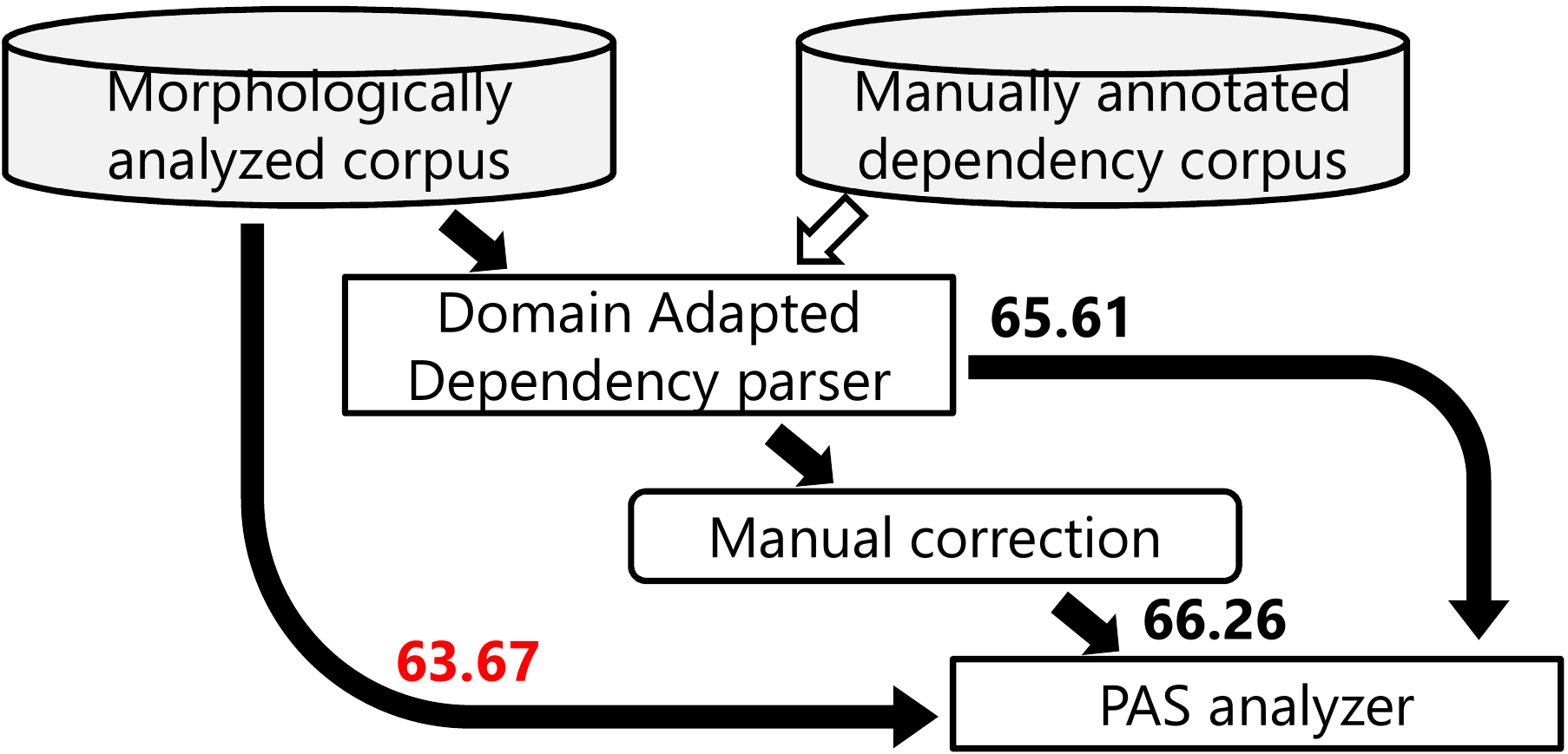}
\caption{Overall conclusion of this work.}
\label{Figure:result}
\end{figure}

\section{Conclusion}
This paper clarified the effect of the dependency information in an PASA by comparing the analyzer that does not use dependency information, an analyzer that uses oracle dependency information, and an analyzer that uses real parsing result including parsing errors.
The result indicated that dependency information improves PASA.
Because analyzers using the dependency information require dependency annotation and parser construction, the cost is very high, which is disproportionate to the improvement in accuracy as shown in {\bf Figure~\ref{Figure:result}}.
The experimental results showed that the indirect features, which compensate for the absence of dependency features, worked well enough.
By considering the cost of preparing dependency information, the accuracy is reasonable to use in the realistic situations.
We plan to design a framework of rapid data preparation, and adopt the system to a variety of domains in the future.

\bibliography{naaclhlt2016}
\bibliographystyle{naaclhlt2016}

\end{document}